\documentclass[10pt,twocolumn,letterpaper]{article}

 \usepackage[pagenumbers]{cvpr} 

\usepackage[dvipsnames]{xcolor}

\definecolor{cvprblue}{rgb}{0.21,0.49,0.74}
\usepackage[pagebackref,breaklinks,colorlinks,citecolor=cvprblue]{hyperref}
\usepackage{graphicx}
\usepackage{amsmath}
\usepackage{amssymb}
\usepackage{booktabs}
\usepackage{times}
\usepackage{soul}
\usepackage{url}
\usepackage{amsmath}
\usepackage{amsthm}
\usepackage{amsfonts}
\usepackage{booktabs}
\usepackage[switch]{lineno}
\usepackage{color}
\usepackage{multirow}
\usepackage{float}
\usepackage{colortbl}
\usepackage{algorithm}
\usepackage{algorithmicx}
\usepackage{algpseudocode}

\def\paperID{12559} 
\def\confName{CVPR}
\def\confYear{2024}

\title{PanBench: Towards High-Resolution and High-Performance Pansharpening}

\author{Shiying Wang$^{1, }$\footnotemark[1], Xuechao Zou$^{1, }$\footnotemark[1], Kai Li$^{2}$, Junliang Xing$^{2}$, Pin Tao$^{1,2, }$\footnotemark[2]\\
$^1$Department of Computer Technology and Applications, Qinghai University, Xining, China\\
$^2$Department of Computer Science and Technology, Tsinghua University, Beijing, China \\
{\tt\small wangshiying.qhu@foxmail.com, xuechaozou@foxmail.com, tsinghua.kaili@gmail.com}, \\ 
{\tt\small jlxing@tsinghua.edu.cn, taopin@tsinghua.edu.cn}}

\begin{document}
\maketitle

\begin{abstract}

Pansharpening, a pivotal task in remote sensing, involves integrating low-resolution multispectral images with high-resolution panchromatic images to synthesize an image that is both high-resolution and retains multispectral information. These pansharpened images enhance precision in land cover classification, change detection, and environmental monitoring within remote sensing data analysis. While deep learning techniques have shown significant success in pansharpening, existing methods often face limitations in their evaluation, focusing on restricted satellite data sources, single scene types, and low-resolution images. This paper addresses this gap by introducing PanBench, a high-resolution multi-scene dataset containing all mainstream satellites and comprising 5,898 pairs of samples. Each pair includes a four-channel (RGB + near-infrared) multispectral image of 256$\times$256 pixels and a mono-channel panchromatic image of 1,024$\times$1,024 pixels. To achieve high-fidelity synthesis, we propose a Cascaded Multiscale Fusion Network (CMFNet) for Pansharpening. Extensive experiments validate the effectiveness of CMFNet. We have released the dataset, source code, and pre-trained models in the supplementary, fostering further research in remote sensing.
\end{abstract}

\section{Introduction}
\label{intro}

With the continuous advancement of remote sensing technology, a substantial amount of earth observation data can be readily obtained. Due to the technological constraints of remote sensing imaging sensors, remote sensing images are typically provided as low-resolution multispectral (MS) images and high-resolution panchromatic (PAN) images. The task of pansharpening~\cite{zhang2021gtp} has emerged to obtain images with high spatial and spectral resolutions. Pansharpening aims to reconstruct a high-resolution MS image from a low-resolution MS image guided by a PAN image.
The pansharpened images enable better analysis and interpretation of the data, serving as an indispensable preprocessing step for downstream tasks in remote sensing, such as land cover classification~\cite{cao2020hyperspectral,lv2023novel}, object recognition~\cite{liu2021sraf,li2023recognizingobject}, and change detection~\cite{asokan2019change,wu2023fully}. It provides high-quality data for downstream remote sensing~\cite{cao2021deep,jiang2018deep,wang2020ultra} applications.

Deep learning methods~\cite{dong2015imageCNN,lecun1995convolutional,shao2019residualGAN,ma2020panGAN} have been widely applied in recent years and have made significant progress in pansharpening. It usually utilizes deep neural network models~\cite{hu2021speech,li2022efficient} to learn the complex mapping relationship between MS and PAN images through end-to-end training. 

Existing deep learning-based pansharpening algorithms for remote sensing images can be broadly categorized into three types: 1) pixel-level fusion, 2) feature-level fusion, and 3) pixel-feature-level fusion. Pixel-level fusion splices the up-sampled MS image and PAN image directly on the channel dimension and then inputs them into the neural model for processing. Such as PNN~\cite{2016PNN}, MSDCNN~\cite{yuan2018MSDCNN}, GPPNN~\cite{xu2021gppnn} and so on. While pixel-level fusion is easy to implement, it may potentially destroy the spectral information of the MS image, resulting in fused images that are not sufficiently clear in certain scenarios. Feature-level fusion is to extract modality-aware features from the panchromatic and multispectral images independently, followed by information fusion in the feature space. Algorithms in this category include PanNet~\cite{yang2017pannet}, TFNet~\cite{liu2020tfnet}, FusionNet~\cite{deng2020detailfusionnet}, among others. This kind of method can better retain the band information of MS images. Still, it needs many transformations and calculations and is easily affected by the feature extraction algorithm. Pixel-feature-level level fusion refers to injecting PAN images as reference information into the network in stages based on image super-resolution tasks to guide the entire spatial information reconstruction process. Such as SFIIN~\cite{zhou2022SFIIN}, MIDP~\cite{zhou2022mutualMIDP}. Some of these algorithms introduce an information-driven framework to reduce redundancy and enhance model performance.

\begin{figure*}
  \centering
  \noindent\includegraphics[width=\linewidth]{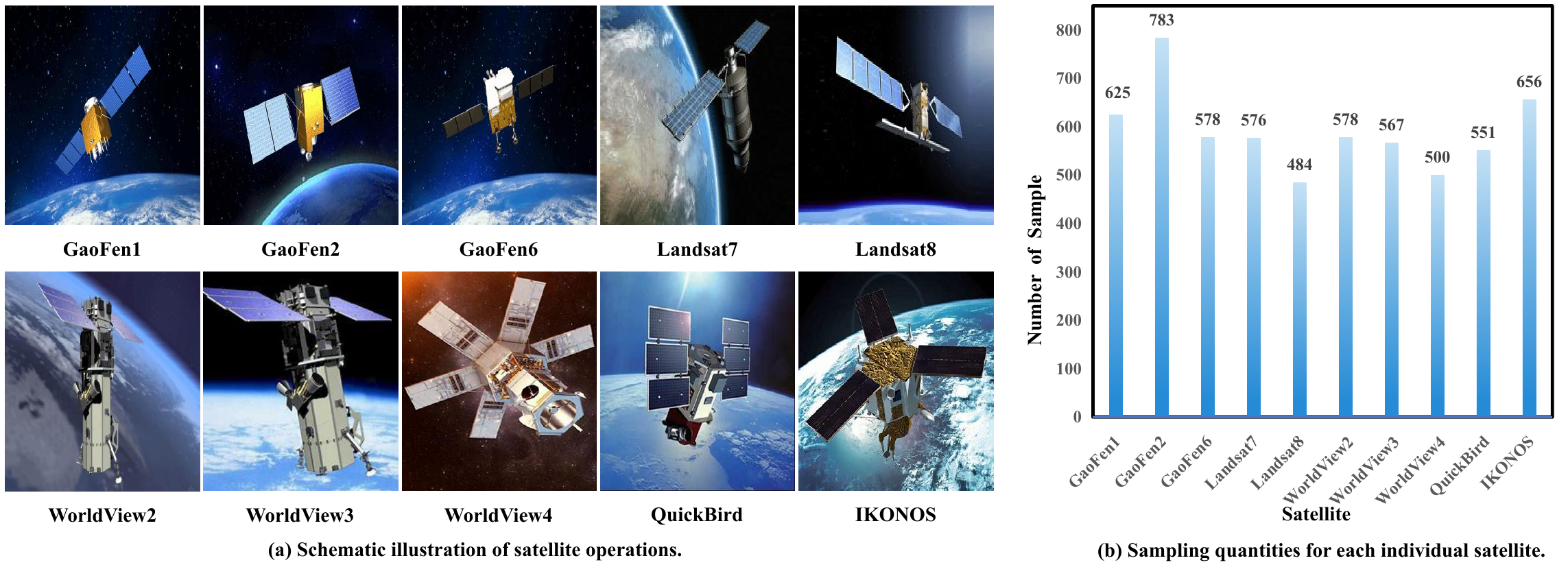}
  \caption{All mainstream satellites for pansharpening are supported by our PanBench dataset. 
  }
  \label{fig: satellite}
\end{figure*}
\begin{table*}[]
\centering
\caption{Comparative analysis of the PanBench dataset against datasets used in other representative literature.}
\label{tab:schedule}
\setlength{\tabcolsep}{3.7pt}
\scalebox{1.00}{
\begin{tabular}{l|c|c|cccccccccc|c}
\toprule
\textbf{Method}    & \textbf{Publication}            & \textbf{Year} & \textbf{GF1} & \textbf{GF2} & \textbf{GF6} & \textbf{LC7} & \textbf{LC8}  & \textbf{WV2} & \textbf{WV3} & \textbf{WV4} & \textbf{QB} & \textbf{IN} & \textbf{PAN} \\ \midrule
PNN~\cite{2016PNN}       & Remote Sens.       & 2016 &     &     &     &     &     & $\checkmark$     &    &    &     & $\checkmark$     & 132$\times$132    \\
PanNet~\cite{yang2017pannet}    & ICCV               & 2017 &     &     &     &     &     & $\checkmark$     & $\checkmark$    &    &    &$\checkmark$     & 400$\times$400  \\
MSDCNN~\cite{yuan2018MSDCNN}    & J-STARS            & 2018 &     &     &     &     &     & $\checkmark$     &     &     & $\checkmark$     & $\checkmark$     & 164$\times$164    \\
TFNet~\cite{liu2020tfnet}     & Inform. Fusion & 2020 & $\checkmark$   &     &     &    &     &     &     &     &$\checkmark$     &     & 512$\times$512  \\
FusionNet~\cite{deng2020detailfusionnet} & TGRS               & 2020 &     & $\checkmark$   &     &    &     & $\checkmark$     & $\checkmark$     &    & $\checkmark$     &     & 64$\times$64    \\
PSGAN~\cite{liu2020psgan}     & TGRS               & 2020 &     & $\checkmark$   &     &    &     & $\checkmark$     &     &     & $\checkmark$     &     & 256$\times$256    \\
GPPNN~\cite{xu2021gppnn}     & CVPR               & 2021 &     & $\checkmark$   &     &    & $\checkmark$     &     &     &     & $\checkmark$     &     & 128$\times$128    \\
SRPPNN~\cite{cai2020srppnn}    & TGRS               & 2021 &     &     &     &    & $\checkmark$     &     & $\checkmark$     &     & $\checkmark$     &     & 256$\times$256    \\
MDSSC-GAN~\cite{gastineau2021generativeMDSSC-GAN} & TGRS               & 2021 &     &     &     &    &     &     & $\checkmark$    &     &     &     & 512$\times$512  \\
MIDP~\cite{zhou2022mutualMIDP}      & CVPR               & 2022 &     & $\checkmark$   &     &    &     & $\checkmark$     & $\checkmark$    &     &     &     & 128$\times$128    \\
SFIIN~\cite{zhou2022SFIIN}     & ECCV               & 2022 &     & $\checkmark$   &     &    &     & $\checkmark$     & $\checkmark$    &     &     &     & 128$\times$128    \\
PanDiff~\cite{meng2023pandiff}   & TGRS               & 2023 &     & $\checkmark$   &     &    &     & $\checkmark$     & $\checkmark$     &     & $\checkmark$     &     & 64$\times$64    \\
USSCNet~\cite{liu2023supervised}   & Inform. Fusion & 2023 &     &     &     &     &     & $\checkmark$     & $\checkmark$    &     &     & $\checkmark$     & 256$\times$256    \\
PGCU~\cite{zhu2023PGCU}      & CVPR               & 2023 &     & $\checkmark$   &     &    &     & $\checkmark$     & $\checkmark$    &     &     &     & 128$\times$128    \\
\rowcolor[RGB]{217,217,217}CMFNet [Ours]     & - &  2023  & $\checkmark$   & $\checkmark$   & $\checkmark$   & $\checkmark$  & $\checkmark$   & $\checkmark$   & $\checkmark$  & $\checkmark$   & $\checkmark$   & $\checkmark$   & 1024$\times$1024  \\ \bottomrule
\end{tabular}}
\end{table*}

Although deep learning-based pansharpening methods for remote sensing images have achieved promising results, numerous challenges and issues remain to be addressed. On the one hand, the current research on pansharpening relies on small-scale patch cropping, limited satellite sources, and a single category of scenes, which can lead to poor overall generalization capability of the models. On the other hand, in the existing methods, MS and PAN are usually concatenated for channel splicing, the feature extraction process is coupled, and the cascade is not considered for multiscale fusion, which indirectly increases the difficulty of image detail recovery. Therefore, a unified high-resolution dataset and stable high-performance baseline are urgently needed for pansharpening to better meet the requirements of practical applications.
We have compiled a more comprehensive dataset, PanBench, to address the issue mentioned above by gathering remote sensing data from multiple satellites. Subsequently, based on this dataset, we introduce the Cascaded Multiscale Fusion Network (CMFNet). Hierarchical image feature extraction is performed on the input PAN and MS images, respectively, and dense cascade and top-down cascade addition are used to improve the interconnection between the features of the decoding layer. The hierarchical feature maps are directly densely connected, and all coarse low-dimensional features are exploited to generate satisfactory high-resolution depth outputs without much attenuation through the decoding layers. Thus, the fused features are decoded to restore the pasharpening image.

\section{Related Work}

\subsection{Datasets for Pansharpening}

The dataset plays a crucial role in developing and evaluating pansharpening algorithms. However, previous research (\cref{tab:schedule}) exhibits issues in the following aspects: 1) The majority of studies employ two to four satellite images to train and validate the effectiveness of pansharpening algorithms. Among them, GaoFen2 (GF2), IKONOS (IN), QuickBird (QB), WorldView2 (WV2), and WorldView3 (WV3) are commonly used. For instance, PanNet~\cite{yang2017pannet} utilizes IN, WV2, and WV3 datasets, while FusionNet~\cite{deng2020detailfusionnet} employs GF2, QB, WV2, and WV3 datasets. However, satellite data sources are relatively limited, and the algorithm's generalization on other satellites needs to be verified. 2) The input multispectral scales are small. For example, PSGAN~\cite{liu2020psgan} and RSIF~\cite{jin2021RSIF} use 64$\times$64 MS images as input, while MIDP~\cite{zhou2022mutualMIDP} and SFIIN~\cite{zhou2022SFIIN} adopt 32$\times$32 MS images as input. It is not competent for batch processing tasks with large-resolution remote sensing images of the real world. 3) The datasets used did not specify land cover categories or included only a single category, such as PGCU~\cite{zhu2023PGCU}, PanDiff~\cite{meng2023pandiff}, and USSCNet~\cite{liu2023supervised}.

This paper constructs PanBench (\cref{fig: satellite}), a large-scale, high-resolution, and multi-scene dataset containing all major satellites used for pansharpening. To ensure the scene's diversity, according to \cite{demir2018challenge,cheng2017remoteClassification}, the scene from PanBench is divided into water, urban, ice/snow, crops, vegetation, and barren. It ensures the algorithm can effectively handle different scenarios and produce reliable results.

\subsection{Algorithms for Pansharpening}\label{sec: alg}

Pansharpening aims to fuse low-resolution MS images with high-resolution PAN images to obtain high-resolution MS images. In recent years, deep learning-based~\cite{huang2015} pansharpening methods have received significant attention, which relies on large-scale data to learn the nonlinear relationship between ideal fused high-resolution MS images and low-resolution MS and PAN images. In general, deep learning-based pansharpening methods take the original MS as the ground truth based on Wald protocol ~\cite{wald1997fusion}, conduct network training under reduced resolution, and apply the trained model to the original PAN and MS images directly to obtain full-resolution fusion images.
Typical pansharpening methods based on deep learning mainly include two kinds of network structures: single-branch and double-branch. For instance, inspired by the success of residual networks~\cite{he2016Resnet}, Yang \etal \cite{yang2017pannet} proposed a deep network called PanNet for pansharpening, which is trained in the high-pass domain to preserve spatial information. The up-sampled MS image is added to the output of the final network through residual connections, thereby preserving spectral information. 
Cai \etal \cite{cai2020srppnn} developed a novel convolutional neural network-based deep super-resolution pansharpening algorithm (SRPPNN) that uses multiscale features in MS images to reconstruct spatial information. Distinguishing the single-branch structure above, two-branch structures usually extract features from PAN and low-resolution MS images, respectively, and then fuse them in hidden space to reconstruct high-resolution MS images. For example, Xu \etal~\cite{xu2021gppnn} proposed a new paradigm combining depth expansion and observation models to develop model-driven pansharpening network GPPNN. This pansharpening model takes PAN and low-resolution MS images into account. However, the above pansharpening methods did not explicitly perform complementary learning of information between PAN and MS images, further limiting the performance. Man Zhou \etal \cite{zhou2022SFIIN} proposes a new mutual information-driven pansharpening framework, which can reduce information redundancy and improve model performance. Zhu \etal~\cite{zhu2023PGCU} has developed a new probability-based global cross-modal up-sampling method for translation sharpening, taking full advantage of the global information of each pixel in low-resolution MS images and the cross-modal information of guided PAN images. However, the existing methods do not consider the cascaded fusion of multi-scale spectral information and spatial information, resulting in unstable image synthesis performance.

\section{PanBench Dataset}

We construct the PanBench, a novel unified evaluation dataset for pansharpening, which supports 10 mainstream satellites, contains 5,898 high-resolution sample pairs, and has 6 manually labeled land cover classifications. This enables us to capture the challenges of complexity and variability encountered in practical generalization applications and contributes to further advancements in this field.

\subsection{Multi-Source Satellite} 

A broader range of data sources is required to develop effective algorithms for pansharpening tasks under various conditions.
The PanBench created includes the primary satellite sources currently available for pansharpening, as shown in \cref{fig: satellite}~(a). Not all satellites support pansharpening, and some satellites, such as Sentinel-2, do not meet pansharpening's conditions because it does not mount sensors that capture the panchromatic band. The commonly used pansharpening in the GaoFen series includes GF1 and GF2. PanBench adds the GF6 data set. The GF1, GF2, and GF6 datasets contain 625, 783, and 578 image pairs, respectively. The Worldview series also supplements the WV4 dataset based on the commonly used WV2 and WV3. The WV2, WV3, and WV4 datasets contain 578, 567, and 500 image pairs, respectively. The Landsat series adds the LC7 dataset (576 image pairs) to the LC8 dataset (484 image pairs). In addition, there are frequently used QB (551 image pairs) and IN (656 image pairs) datasets(\cref{fig: satellite}~(b)). Thereby, the model can learn the characteristics and rules of satellites more comprehensively and accurately and improve the robustness of the model.

\subsection{Pre-Processing} Loading the whole large remote sensing image may occupy a lot of computing resources and time. The data size can be reduced and the efficiency of data processing can be improved by clipping the image. Firstly, we performed a series of pre-processing operations on the collected source data: radiometric calibration~\cite{lin2004radiometric}, atmospheric correction~\cite{liang2001atmospheric}, orthometric correction~\cite{barzaghi2023gravity} and image alignment to initially extract and enhance useful information in the images. Secondly, we segmented the entire large-scale remote sensing image into some image pairs, which were composed of four-channel (RGB + near-infrared) MS images with a $256\times 256$ pixel size and single-channel PAN images with a $1,024\times 1,024$ pixel size. The clipping scale is the largest compared with the training data set in the current representative literature. The larger the image size, the more detail, context information, and pixel information can be captured, helping to accurately capture the subtle features and structure in the image. For Landsat7 and Landsat8, the spatial resolution of the MS component is half that of the PAN. In other cases, the MS component's spatial resolution is a quarter of the PAN's. Ultimately, PanBench comprises 5,898 image pairs.

\subsection{Scene Classification} 
\begin{figure}
  \centering
  \noindent\includegraphics[width=1.0\linewidth]{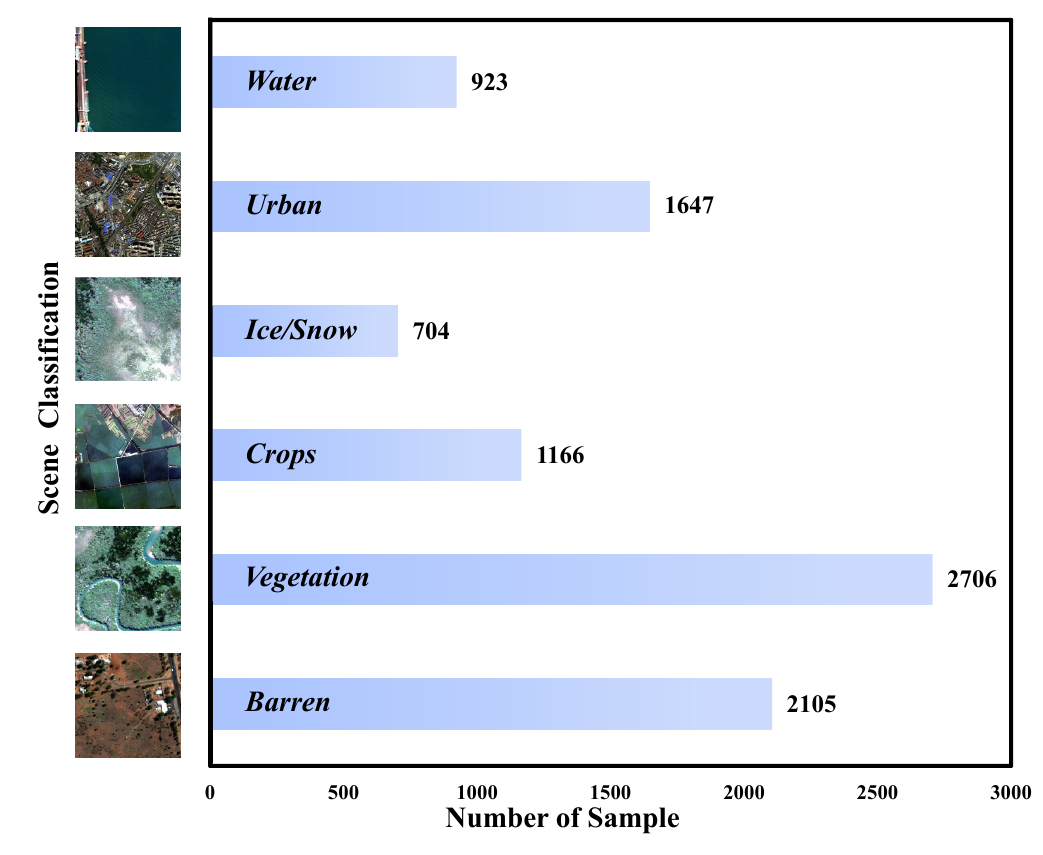}
  \caption{Number of six scenes of land cover classification.}
  \label{fig: scene}
\end{figure}

The ultimate objective of pansharpening is to use fused images to provide crucial spatial information for downstream tasks such as urban planning, land management, disaster response, and more. Therefore, it is necessary to verify the accuracy of the pansharpening method in various scenarios. In this study, by labeling the training data of different scenarios, we reclassified the data into six basic land cover types according to DeepGlobe 2018~\cite{demir2018challenge} and Cheng \etal~\cite{cheng2017remoteClassification}, Meng \etal~\cite{meng2022multilayerClassification}: water, urban, ice/snow, crops, vegetation, and barren. The specific sample counts are shown in \cref{fig: scene}. The algorithm's performance in different scenes may differ, so optimizing the model's parameters and the algorithm design is necessary by verifying and evaluating different scenes.

\section{Method}
\begin{figure*}
  \centering
  \noindent\includegraphics[width=1.0\linewidth]{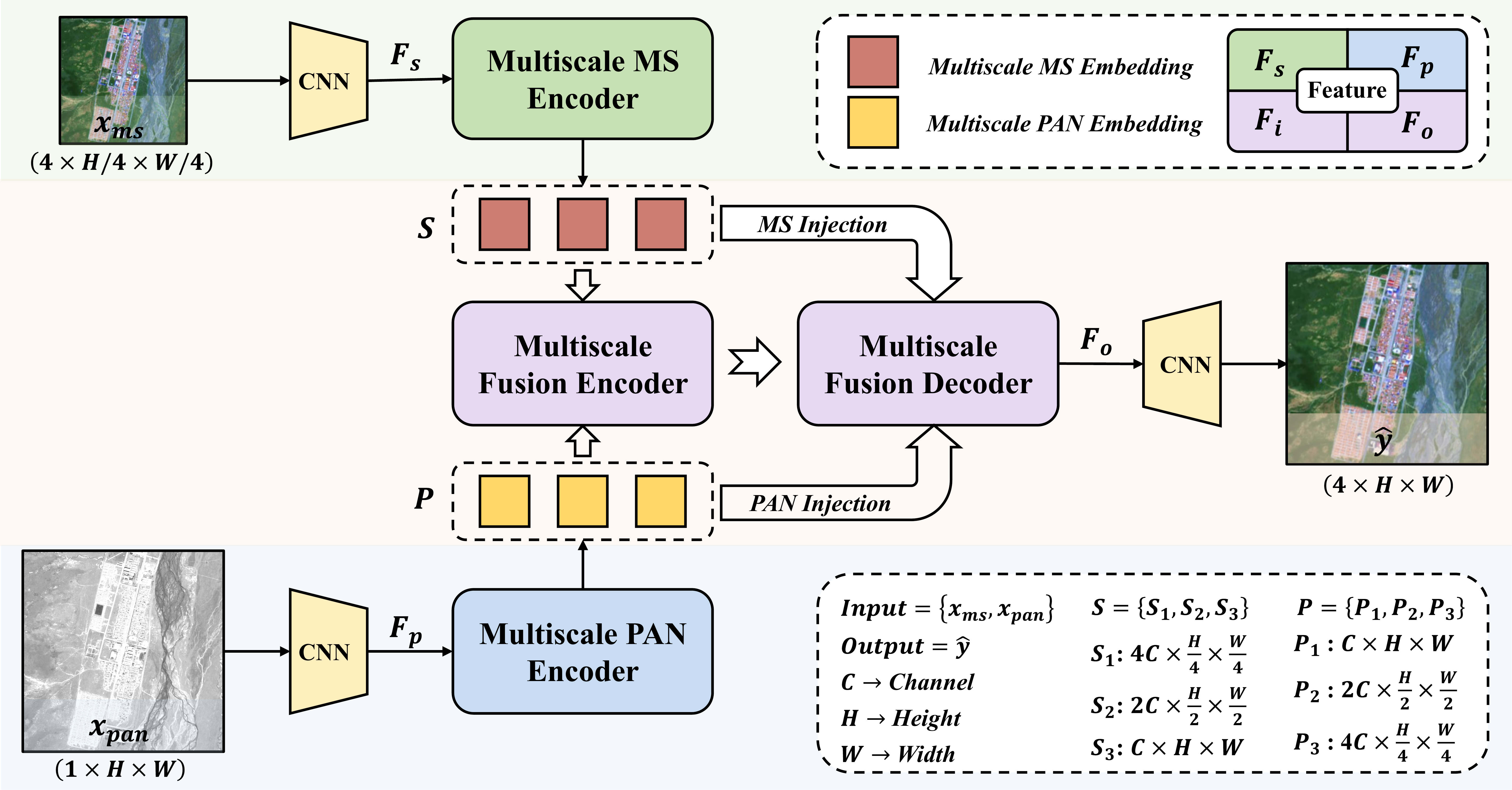}
  \caption{ The overall framework of our proposed CMFNet, where $\mathbf{F}_i$ and $\mathbf{F}_o$ represent the input and output of the autoencoder, respectively.}
  \label{fig: framework}
\end{figure*}

\begin{figure*}
  \centering
  \noindent\includegraphics[width=1.0\linewidth]{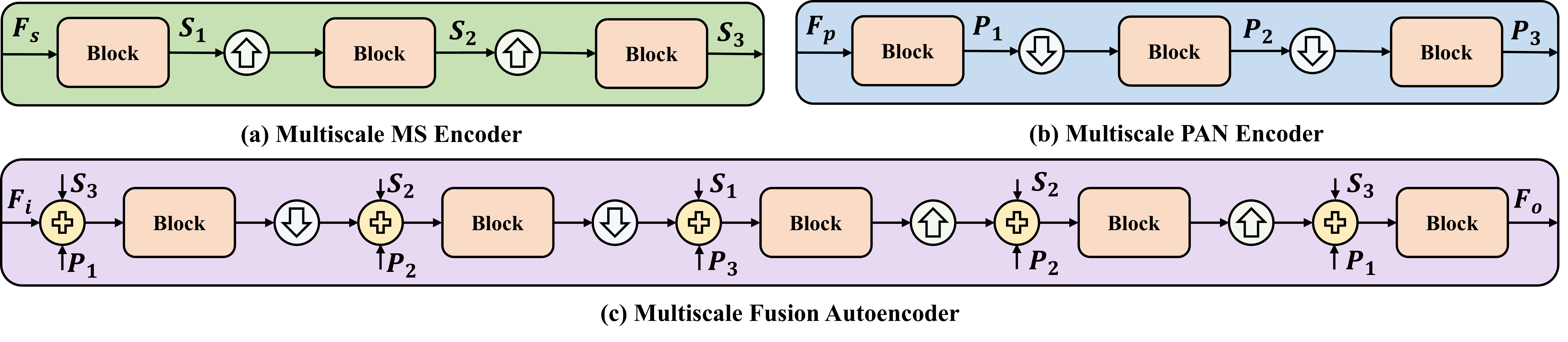}
  \caption{The detailed structure of three components in the CMFNet: (a) Multiscale MS encoder, (b) Multiscale PAN encoder, (c) Multiscale fusion autoencoder. The upward and downward arrows indicate upsampling and downsampling, respectively, while the plus sign denotes summation. The term ``block'' can be substituted by any module, such as ResNet \cite{resnet}, NAFNet \cite{nafnet}, ConvNeXt \cite{convnext} block, \etc. The autoencoder integrates spectral and spatial information by incorporating identical-scale features from injecting two encoders.}
  \label{fig: model}
\end{figure*}

\subsection{Overall Framework} \label{sec:framework}
\subsubsection{Problem Definition}
The purpose of the pansharpening is to fuse a low-resolution MS image $\mathbf{x_{ms}}\in \mathbb{R}^{4\times \frac{H}{4}\times \frac{W}{4}}$ with a high-resolution PAN image $\mathbf{x_{pan}}\in \mathbb{R}^{1\times H\times W}$ to produce a high-resolution color image $\mathbf{y}\in \mathbb{R}^{4\times H\times W}$ with multispectral information, where C, H and W represent the channels, height and width of the image, respectively. Pansharpening's task flow can be expressed in the following formula:
\begin{equation}
    \mathbf y={\boldsymbol{f}_{\theta}}(x_{ms},x_{pan}),
\end{equation}
where $\boldsymbol{f}_{\theta}$ is a parameterized neural network model.

\subsubsection{Overall Pipeline} 

In this paper, we proposed the CMFNet, a high-fidelity fusion network for pansharpening. It primarily consists of three components: the multiscale MS encoder, the multiscale PAN encoder, and the multiscale fusion autoencoder. 

The overall pipeline of the CMFNet is illustrated in \cref{fig: framework}, which is described as follows:
\begin{itemize}
    \item \textbf{Multiscale MS Encoder.} The MS image $x_{ms}$ passes through a convolutional layer with a kernel size of 3$\times$3 to obtain the features $\mathbf{F}_s\in \mathbb{R}^{4C\times \frac{H}{4}\times \frac{W}{4}}$. The features are passed through a multiscale MS encoder, obtaining hierarchical image features of three resolutions.
    \item \textbf{Multiscale PAN Encoder.} Similar to the multiscale MS encoder, the PAN features $\mathbf{F}_p\in \mathbb{R}^{C\times H\times W}$ are pre-extracted without changing the image size, and three features corresponding to the MS scale are obtained by the multiscale PAN encoder.
    \item \textbf{Multiscale Fusion Autoencoder.} The fused features $\mathbf{F}_o\in \mathbb{R}^{C\times H\times W}$ are obtained from the features of the three corresponding scales obtained from the multiscale MS encoder and the PAN encoder, respectively. Finally, the fused features are output by a convolution layer.
\end{itemize}

\subsection{Cascaded Multiscale Fusion Network} \label{sec:network}

\subsubsection{Multiscale MS Encoder}
Ground objects and scenes have different characteristics at different scales due to their various sizes and shapes. Through multiscale feature extraction, the pansharpening network can focus on local details and global structure at the same time, to better capture the subtle changes and context information of ground objects. We introduce the multiscale MS encoder module within the CMFNet network, as shown in \cref{fig: model} (a). The features $\mathbf F_{s}$ of the MS are processed through three blocks \cite{zou2023diffcr} and two upsampling operations, yielding three features at distinct resolutions, denoted as $\mathbf S=\left \{ \mathbf S_{i}\in \mathbb{R}^{(C\times 2^{3-i})\times \frac{H}{2^{(3-i)}}\times \frac{W}{2^{(3-i)}}}| i\in[1,2,3]\right \}$, where the upsampling operation denotes the interpolation method.

\subsubsection{Multiscale PAN Decoder}

PAN images have higher spatial resolution than MS images. Multiscale feature extraction of PAN images can compensate for the lack of spatial perception of MS images and capture a wide range of structures and spatial relationships from PAN images to better guide the pansharpening process and improve the clarity and quality of pansharpening results. Therefore, we design the multiscale PAN encoder module in the CMFNet network, as shown in \cref{fig: model} (b). For the PAN image feature $\mathbf{F}_p$, three features corresponding to the scale of MS are obtained by three blocks, same as MS encoder and two downsampling operations, that is, $\mathbf P=\left \{ \mathbf P_{i}\in \mathbb{R}^{(C\times 2^{i-1})\times \frac{H}{2^{(i-1)}}\times \frac{W}{2^{(i-1)}}}| i\in[1,2,3]\right \}$, where the downsampling operation denotes the pooling layer.

\subsubsection{Multiscale Fusion Autoencoder}
\paragraph{Multiscale Fusion Encoder.}

The encoder is responsible for gradually downsampling the input image and extracting high-level semantic features. To be able to make full use of the information of the images of the multiscale MS encoder and multiscale PAN encoder, it is necessary to deeply fuse $\mathbf{S}$ and $\mathbf{P}$ at the same scale. After multiscale MS and PAN encoders, we have two feature sets $\mathbf{S}$ and $\mathbf{P}$, representing PAN and MS images, respectively. Since high-resolution MS images must have high spatial and spectral resolutions, their features must have both spatial and spectral information. To do this, the two feature sets must be concatenated and added at the same scale. 

That is, $\mathbf F_{1}$=$\mathbf S_{3}$ + $\mathbf P_{1}$, $\mathbf F_{2}$=$\mathbf S_{2}$ + $\mathbf P_{2}$, and $\mathbf F_{3}$=$\mathbf S_{1}$ + $\mathbf P_{3}$. Then the block same as the multiscale MS encoder is used to encode the concatenated feature maps into a more compact representation after each addition, and the end of the multiscale fusion encoder (\cref{fig: model} (c)) is the feature set $\mathbf E=\left \{ \mathbf E_{i}\in \mathbb{R}^{(C\times 2^{i-1})\times \frac{H}{2^{(i-1)}}\times \frac{W}{2^{(i-1)}}}| i\in[1,2,3]\right \}$, which encodes the spatial and spectral information of the two input images,

\begin{equation}
    \mathbf{E}_1 = \operatorname{Block}(\mathbf{F}_i),
\end{equation}
\begin{equation}
    \mathbf{E}_i = \operatorname{Block}(\mathbf S_{4-i}+\mathbf{P}_i+(\operatorname{Downsample}(\mathbf{\mathbf E_{i-1}})).
\end{equation}

\paragraph{Multiscale Fusion Decoder.}

The decoder (\cref{fig: model} (c)) corresponds to the encoder, and the upsampled feature map is fused with the feature map in the corresponding encoder by the feature fusion operation. This can help to recover the detail and texture information of the image.

Specifically, we upsample the features of the set $\mathbf{E}$ and superimpose and fuse them with the features of the corresponding scale of $\mathbf{E}$. In encoder downsampling, some details and local information may be lost due to the loss of information or resolution degradation caused by downsampling. 

Therefore, in the process of decoding $\mathbf{E}$, multiscale injection of $\mathbf{S}$ and $\mathbf{P}$ can obtain the details and local information from the encoder in the decoder, which can effectively connect and fuse the low-level and high-level features,

and finally output the fusion result $\mathbf{F}_o$.

\section{Experiments}

\subsection{Implementation Details}

\paragraph{Training Settings.} The dataset is divided into training, validation, and test sets in a ratio of 8:1:1, which means the training set comprises 4,718 image pairs, and both the validation and test sets contain 590 image pairs each. The batch size is set to 16. We used the Adam~\cite{adam} optimizer with a learning rate of 0.002. If there is no improvement in the performance of the validation set for ten consecutive epochs, the learning rate is reduced to 10\% . 
Cascaded layers in the multiscale feature fusion are set to 3, and the initial number of channels for the CMFNet is 32. We used the MSE as the object function. The training was performed on four NVIDIA GeForce RTX 3090 GPUs.

\paragraph{Evaluation Metrics.} The evaluation of pansharpening algorithms for PAN and MS image fusion involves adapting widely used image quality assessment (IQA) metrics. These metrics include peak signal-to-noise ratio (PSNR), structural similarity index (SSIM) ~\cite{ssim}, spectral angle mapper (SAM)~\cite{SAM}, relative dimensionless global error synthesis (ERGAS)~\cite{ERGAS}, spatial correlation coefficient (SCC)~\cite{SCC}, and mean-square error (MSE, $\times$$10^{-4}$). These metrics measure fidelity, similarity, spectral and spatial distortion, and spatial correlation. By employing these metrics, the performance of pansharpening algorithms can be objectively evaluated and compared in real-world scenarios. This evaluation framework enables comprehensive analysis and comparison of different methods for pansharpening.

\subsection{Comparison with State-of-the-Arts}
\begin{table*}[t]
\centering
\setlength{\tabcolsep}{15pt}
\caption{Quantitative metrics for all the comparison methods on the PanBench.}
\begin{tabular}{l|lllllc}
\toprule
Method & PSNR~$\uparrow$    & SSIM~$\uparrow$   & SAM~$\downarrow$    & ERGAS~$\downarrow$  & SCC~$\uparrow$    & MSE~$\downarrow$     \\ \midrule
PNN~\cite{2016PNN}                                                   & 28.9029 & 0.7887 & 0.0750 & 4.4998 & 0.8992 & 25.5223 \\
PanNet~\cite{yang2017pannet}                                                & 30.1465 & 0.8497 & 0.0702 & 3.8053 & 0.9234 & 18.5474 \\
MSDCNN~\cite{yuan2018MSDCNN}                                                & 29.2675 & 0.8237 & 0.0761 & 4.1677 & 0.9139 & 21.7871 \\
TFNet~\cite{liu2020tfnet}                                                 & 32.7018 & 0.8931 & 0.0600 & 2.8816 & 0.9486 & 11.9169 \\
FusionNet~\cite{deng2020detailfusionnet}                                             & 24.4378 & 0.7175 & 0.0880 & 7.5029 & 0.7913 & 70.0308 \\
GPPNN~\cite{xu2021gppnn}                                                 & 28.8901 & 0.8211 & 0.0842 & 4.2720 & 0.9124 & 22.3945 \\
SRPPNN~\cite{cai2020srppnn}                                                & 31.3186 & 0.8647 & 0.0666 & 3.3526 & 0.9344 & 15.5632 \\
PGCU~\cite{zhu2023PGCU}                                                  & 29.9692 & 0.8244 & 0.0759 & 3.9113 & 0.9167 & 19.5983 \\
\rowcolor[RGB]{217,217,217} CMFNet {[}Ours{]}                                        & \textbf{34.3852} & \textbf{0.9139} &\textbf{0.0579} & \textbf{2.7846} & \textbf{0.9494} & \textbf{9.0428}  \\ \bottomrule
\end{tabular}
\label{tab:all_results}
\end{table*}

\begin{table}[t]
\centering
\caption{Evaluation of CMFNet on each satellite of PanBench.}
\label{tab: satellite}
\setlength{\tabcolsep}{0.8pt}
\begin{tabular}{c|cccccc}
\toprule
Satellite
                                   & PSNR~$\uparrow$  & SSIM~$\uparrow$  & SAM~$\downarrow$     & ERGAS~$\downarrow$  & SCC~$\uparrow$    & MSE~$\downarrow$ \\ \midrule
GF1                                &42.8659       & 0.9591      & 0.0196  & 1.0457 & 0.9580 &1.9661     \\
GF2                                &36.6737       &0.9554       & 0.0394  & 2.2247 & 0.9870 &3.0906     \\
GF6                                &34.2858       &0.9165       & 0.0361  & 1.5311 & 0.9826 &3.9269     \\
LC7                                &36.6354       &0.9391       & 0.0110  & 0.8417 & 0.9918 &2.3282     \\
LC8                                &30.0631       &0.9055       & 0.0719  & 2.6224 & 0.9685 &10.1524     \\
WV2                                &34.8766       &0.9424       & 0.0577  & 2.3661 & 0.9839 &3.8216     \\
WV3                                &35.2992       &0.9350       & 0.0682  & 4.0791 & 0.8546 &10.1377     \\
WV4                                &29.7377       &0.8751       & 0.0894  & 3.4016 & 0.9796 &13.3450     \\
QB                                 &37.4593       &0.9581       & 0.0506  & 1.4965 & 0.9897 &2.0394     \\
IN                                 &25.1965       &0.7549       & 0.0836  & 4.7192 & 0.8960 &39.0390     \\ \bottomrule
\end{tabular}
\end{table}

\begin{table*}[t]
\centering
\caption{Evaluation of CMFNet on scene classification of PanBench.}
\setlength{\tabcolsep}{15pt}
\begin{tabular}{c|lllllc}
\toprule
Scene & PSNR~$\uparrow$    & SSIM~$\uparrow$   & SAM~$\downarrow$    & ERGAS~$\downarrow$  & SCC~$\uparrow$    & MSE~$\downarrow$     \\ \midrule
Water                           & 41.7804 & 0.9453 & 0.0363 & 2.1097 & 0.8990 & 4.3999  \\
Urban                           & 32.4035 & 0.9299 & 0.0632 & 2.8182 & 0.9807 & 8.5807  \\
Ice/snow                        & 34.0811 & 0.9290 & 0.0342 & 1.5132 & 0.9889 & 5.0679  \\
Crops                           & 35.7273 & 0.9449 & 0.0464 & 1.9383 & 0.9866 & 3.3005  \\
Vegetation                      & 32.4267 & 0.9045 & 0.0578 & 2.6551 & 0.9683 & 10.6414 \\
Barren                          & 32.9757 & 0.8848 & 0.0447 & 2.1215 & 0.9632 & 11.4405 \\ \bottomrule
\end{tabular}
\label{tab:scene_results}
\end{table*}

\begin{figure*}
  \centering
  \noindent\includegraphics[width=0.95\linewidth]{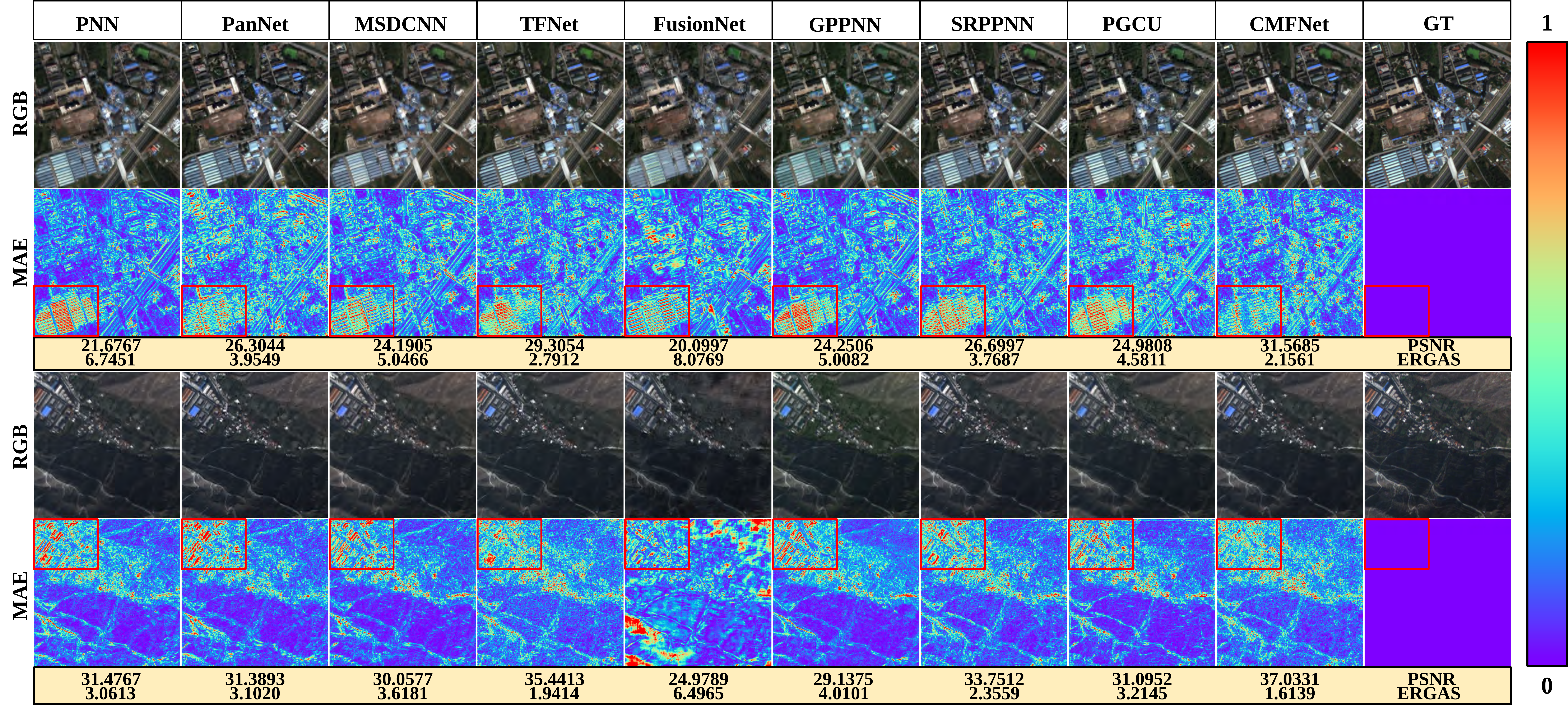}
  \caption{ Visual results generated by different pansharpening methods. MAE denotes the mean absolute error across all spectral bands.
}
  \label{fig: vis}
\end{figure*}

\begin{table}[t]
\centering
\caption{Impact of cascaded layers in the multiscale feature fusion.}\label{tab: cascade}
\setlength{\tabcolsep}{0.5pt}
\begin{tabular}{c|cccccc}
\toprule
\multicolumn{1}{c|}{{\textbf{Cascade}}} & PSNR~$\uparrow$    & SSIM~$\uparrow$   & SAM~$\downarrow$    & ERGAS~$\downarrow$  & SCC~$\uparrow$    & MSE~$\downarrow$     \\ \midrule
1 & 33.0352 & 0.8937 & 0.0579 & 2.7846 & 0.9494 & 10.9749 \\ \midrule
2 & 34.3852 & 0.9139 & 0.0516 & 2.4266 & 0.9595 & 9.0428 \\ \midrule
\rowcolor[RGB]{217,217,217} \textbf{3} & \textbf{34.4921} & \textbf{0.9153} &\textbf{0.0511} & \textbf{2.3984} & \textbf{0.9601} & \textbf{8.8862}  \\ \bottomrule
\end{tabular}
\end{table}

\begin{table}[t]
\centering
\caption{Ablation study results on cascaded information injection.}\label{tab: injection}
\setlength{\tabcolsep}{0.5pt}
\begin{tabular}{c|cccccc}
\toprule
\multicolumn{1}{c|}{\textbf{Injection}} & PSNR~$\uparrow$    & SSIM~$\uparrow$   & SAM~$\downarrow$    & ERGAS~$\downarrow$  & SCC~$\uparrow$    & MSE~$\downarrow$     \\ \midrule
$\times$ & 34.0757 & 0.9102 & 0.0528 & 2.4926 & 0.9568 & 9.5769 \\ \midrule
\rowcolor[RGB]{217,217,217} \textbf{\checkmark} & \textbf{34.4921} & \textbf{0.9153} &\textbf{0.0511} & \textbf{2.3984} & \textbf{0.9601} & \textbf{8.8862}  \\ \bottomrule
\end{tabular}
\end{table}

\begin{table}[h]
\centering
\small
\caption{Ablation study results on the scalability of PanBench dataset. Here, ``SR'', ``CZ'', and ``PS'' represent image super-resolution, image colorization, and pansharpening, respectively.}\label{tab: scalability}
\setlength{\tabcolsep}{0.5pt}
\begin{tabular}{ccccccc}
\toprule
\textbf{Task} & PSNR~$\uparrow$ & SSIM~$\uparrow$ & SAM~$\downarrow$ & ERGAS~$\downarrow$ & SCC~$\uparrow$ & MSE~$\downarrow$ \\ \midrule
SR (w/o PAN) & 29.6283 & 0.7656 & 0.3256 & 25.9604 & 0.8205 & 25.3851\\
CO (w/o MS) & 25.1988 & 0.7811 & 0.1686 & 7.9895 & 0.7782 & 50.1688\\
PS (MS+PAN) & 34.3852 & 0.9139 &0.0579 & 2.7846 & 0.9494 & 9.0428 \\ \bottomrule
\end{tabular}
\end{table}

To assess the performance of CMFNet in the task of pansharpening, we have selected eight representative state-of-the-art (SOTA) methods from 2016 to 2023 for comparative analysis. These methods include PNN~\cite{2016PNN}, Pannet~\cite{yang2017pannet}, MSDCNN~\cite{yuan2018MSDCNN}, TFNet~\cite{liu2020tfnet}, FusionNet~\cite{deng2020detailfusionnet}, GPPNN~\cite{xu2021gppnn}, SRPPNN~\cite{cai2020srppnn}, and PGCU~\cite{zhu2023PGCU}. We conducted comprehensive experiments on the PanBench to evaluate the performance of our model. Then we performed experiments on various classification scenarios to ensure the generalizability of our model across diverse real-world contexts.

\paragraph{Quantitative Comparison.} The comparative results of the nine algorithms on the PanBench are presented in \cref{tab:all_results}, with the best values highlighted in bold black. Our proposed method achieves the best overall performance compared to other state-of-the-art pansharpening methods, firmly establishing its superiority. Specifically, in terms of PSNR, our method outperforms the closest competitor, TFNet, by approximately 1.67. In addition to PSNR, notable improvements can also be observed in other metrics, indicating reduced spectral distortion and preserved spatial textures. We also counted the index values of CMFNet in each scenario(\cref{tab:scene_results}) and satellite (\cref{tab: satellite}). It can be seen that the performance of each metric is optimal due to the simple structure of the water. The evaluation metrics vary from satellite to satellite, which is consistent with objective facts.
\paragraph{Qualitative Comparison.} We also present qualitative comparisons of visualization results (as shown in \cref{fig: vis}) to demonstrate the effectiveness of our method on representative samples from across the datasets. The images in the second and fourth rows are the difference MAE between the output pansharpening results and the ground truth, where the red color means the generated effect is worse, and the blue color means the generated effect is better. Our model has smaller spatial and spectral distortions compared to other competing methods. For MAE poor, we notice that our proposed method is closer to the ground truth than the other compared methods. This result proves the effectiveness of the proposed CMFNet, which successfully reduces information redundancy and enhances the quality of pansharpening results.

\subsection{Ablation Studies}\label{sec:ablation}

\subsubsection{Impact of the Multiscale Cascading}
By two decoupled MS and  PAN encoders, we can obtain three hierarchical multiscale features with dimensions of 64$\times$64, 128$\times$128, and 256$\times$256. The autoencoder seamlessly fuses the corresponding features of the same scale in a cascading manner. As depicted in~\cref{tab: cascade}, we conducted experiments to investigate the impact of various cascading levels on the fusion outcomes. Notably, the performance of pansharpening exhibits linear growth as the number of cascading layers increases, reaching its optimal state when the number of cascading layers equals three. This compellingly validates the necessity of hierarchically cascading fusion.

\subsubsection{Impact of the Cascaded Injection}
In the multiscale fusion decoder, we incorporate a cascading strategy for information injection, whereby the spectral information from the MS image and the spatial information from the PAN image are directly injected across the encoder into the decoder. To validate the effectiveness of this strategy, we conducted ablative research, and the experimental results are shown in~\cref{tab: injection}. Following the introduction of cascading information injection, the PSNR increased by 0.4164, and the ERGAS decreased by approximately 0.1. The consistent performance improvements on all metrics provide substantial evidence for the effectiveness of this strategy. One possible explanation is that for the computer vision task of pansharpening, which belongs to low-level visual computations, the cross-connection of feature injection serves as an efficacious means to alleviate the loss of information during the encoding process.

\subsubsection{Scalability of the PanBench Dataset}
It is worth mentioning that the PanBench dataset we constructed is not only suitable for pansharpening tasks in the field of remote sensing, but also supports other general computer vision tasks such as image super-resolution~\cite{huang2019single,sr3,li2020survey}, image colorization~\cite{palette}, and image classification. In addition, we evaluated the performance of CMFNet on image super-resolution and image colorization tasks through ablation experiments, as shown in \cref{tab: scalability}. Compared to pansharpening, super-resolution~\cite{zhou2023memory} reconstruction only receives PAN images as input, while colorization only receives MS images as input. As can be seen from \cref{tab: scalability}, colorization is the most challenging task, followed by super-resolution and then pansharpening.

\section{Conclusion}

This paper presents PanBench, a large-scale, high-resolution, and multi-scene dataset encompassing the prominent satellites commonly used for pansharpening. The dataset is made available in an open-source manner, to facilitate the development of novel pansharpening methods. To achieve high-fidelity synthesis, we propose CMFNet, a new model designed explicitly for panchromatic sharpening. Experimental results on visual restoration and semantic recovery quality demonstrate the effectiveness of the proposed approach, surpassing existing representative pansharpening methods. The experimental findings also indicate the algorithm's strong generalization capability and impressive performance. We firmly believe that the PanBench dataset will benefit the community, while our evaluations provide valuable directions for future research endeavors.





{
    \small
    \bibliographystyle{ieeenat_fullname}
    \bibliography{main}
}



\end{document}